%% file: iclr2026_conference.tex
\documentclass{article} 
\usepackage{iclr2026_conference,times}

\input{math_commands.tex}

\usepackage{booktabs}
\usepackage{multirow}
\usepackage{colortbl}
\usepackage{hyperref}
\usepackage{url}
\usepackage{graphicx} 
\usepackage{amssymb}
\usepackage{enumitem} 
\usepackage{xcolor}   
\usepackage{siunitx}  
\usepackage{xfp}      
\usepackage[most]{tcolorbox} 
\definecolor{headergray}{RGB}{242,242,242} 
\definecolor{highlightyellow}{RGB}{250,248,245} 
\definecolor{upbg}{RGB}{225,245,225} 
\definecolor{downbg}{RGB}{255,230,230} 
\definecolor{textgreen}{RGB}{30,120,30} 
\definecolor{textred}{RGB}{180,30,30} 

\newcommand{\DeltaSup}[1]{%
  \edef\dval{\fpeval{round(#1,1)}}
  \ifdim \dval pt = 0pt
  \else
    \ifdim \dval pt > 0pt
      \textsuperscript{\colorbox{upbg}{\textcolor{textgreen}{\bfseries\scriptsize$\uparrow$\num{\dval}}}}%
    \else
      \textsuperscript{\colorbox{downbg}{\textcolor{textred}{\bfseries\scriptsize$\downarrow$\num{\fpeval{abs(\dval)}}}}}%
    \fi
  \fi
}

\newcommand{\Score}[2]{%
  \num{#2}\DeltaSup{\fpeval{#2-(#1)}}%
}

\definecolor{auditbg}{RGB}{240, 245, 250}   
\definecolor{auditborder}{RGB}{71, 85, 105} 

\newtcolorbox{AuditProtocol}{
    colback=auditbg,      
    colframe=auditborder, 
    boxrule=0.8pt,        
    arc=2pt,              
    left=15pt, right=15pt, top=15pt, bottom=15pt, 
    enhanced,             
    breakable             
}

\title{GRIP: Geometric Refinement and Adaptive Information Potential for Data Efficiency}

\author{
  Changhao Wang\thanks{Equal contribution} \\
  Politecnico di Torino \\
  \texttt{changhao.wang@polito.it}
  \And
  Jiaolong Yang\textsuperscript{*} \\
  Ant Group
  \And
  Xinhao Yao \\
  Renmin University of China
  \AND 
  Yunfei Yu \\
  Ant Group
  \And 
  Peng Jiao \\
  Ant Group
  \And 
  Lu Yu \\
  Ant Group
  \And
  Junpeng Fang \\
  Ant Group
  \AND 
  Riccardo Cantoro \\
  Politecnico di Torino
  \And
  Qing Cui \\
  Ant Group
  \And
  Jun Zhou \\
  Ant Group
}

\iclrfinalcopy 
\begin{document}

\maketitle
\begin{abstract}
The performance of Large Language Models (LLMs) is increasingly governed by data efficiency rather than raw scaling volume. However, existing selection methods often decouple global distribution balancing from local instance selection, compromising the hierarchical integrity of the training set. We introduce \textbf{GRIP} (Geometric Refinement and Adaptive Information Potential), a framework that unifies these dimensions by modeling the corpus as an information-dense geometric space. GRIP employs a \textbf{Rapid Adaptation Probe (RAP)} to quantify the information potential of semantic clusters, dynamically re-allocating the sampling budget to regions with the highest representation deficits. Subsequently, we perform Intra-Cluster Selection using a \textbf{length-rectified geometric prior} to counteract embedding density artifacts and preserve long-tail logical sequences. Extensive evaluations on Mixture-of-Experts (MoE) models up to 300B tokens demonstrate that GRIP consistently outperforms state-of-the-art baselines, \textbf{surpassing the performance of models trained on $3\times$ larger uncurated datasets}. Our work establishes a robust geometric foundation for adaptive data curation in large-scale pre-training.
\end{abstract}

\section{Introduction}

The generalization of Large Language Models (LLMs) has traditionally relied on the simultaneous scaling of model parameters and training data volume \citep{kaplan2020scaling}. However, current scaling laws suggest that the primary bottleneck for performance has shifted from raw quantity to \textbf{data quality} \citep{hoffmann2022chinchilla, gadre2024language}. As high-quality public corpora approach depletion \citep{villalobos2022will}, simply aggregating noisy web-scale data yields diminishing returns and introduces significant computational waste \citep{gunasekar2023textbooks, goyal2024scaling}. This shift necessitates a rigorous focus on \textbf{data efficiency}: identifying optimal subsets that maximize information gain per unit of compute \citep{li2023starcoder, sorscher2022beyond}.

Recent advances in data efficiency primarily diverge into two paradigms: \textbf{structural budgeting} and \textbf{instance-level saliency}.
The former attempts to balance representational capacity by adjusting mixture weights across predefined domains \citep{xie2023doremi, liu2025quadmix}, yet such coarse-grained adjustments often overlook the underlying semantic cluster and the varying quality within individual clusters \citep{diao2025climb, wang2025can}.
In parallel, instance-centric strategies filter data based on difficulty or training dynamics \citep{zhang2025d3, wang2024capturing}, but these methods frequently decouple local importance from global topology, inadvertently disrupting the structural coherence essential for complex reasoning.
This fragmentation creates a fundamental trade-off: current curators either optimize cluster-level proportions while ignoring instance quality, or filter samples while \textbf{sacrificing the hierarchical integrity} of the corpus.

This structural integrity is particularly vital for \textbf{code corpora}, which form a rigid logical topology defined by brittle syntax and deep hierarchical dependencies \citep{li2023starcoder, guo2020graphcodebert}. In code, the loss of rare but structurally critical snippets can degrade the model’s ability to generalize across logical transitions. Furthermore, Transformer embeddings often suffer from a \textit{geometric collapse} in long-context sequences, where high-information samples are suppressed by \textbf{density artifacts}, making them difficult to identify via standard filters~\cite{ethayarajh2019contextual,zhou2025length}.

To bridge this gap, we introduce \textbf{GRIP} (Geometric Refinement and Adaptive Information Potential), which \textbf{unifies inter-cluster budgeting and intra-cluster selection} within an information-dense geometric space.
The framework first employs a Rapid Adaptation Probe (RAP) to dynamically re-allocate budgets toward semantic clusters with the highest representation deficits.
Subsequently, a \textbf{length-rectified geometric prior} is applied for intra-cluster refinement, preserving global structural variance while capturing high-value, long-tail logical sequences to maximize epistemic gain.

\begin{figure*}[t]
\centering
\includegraphics[width=1\linewidth]{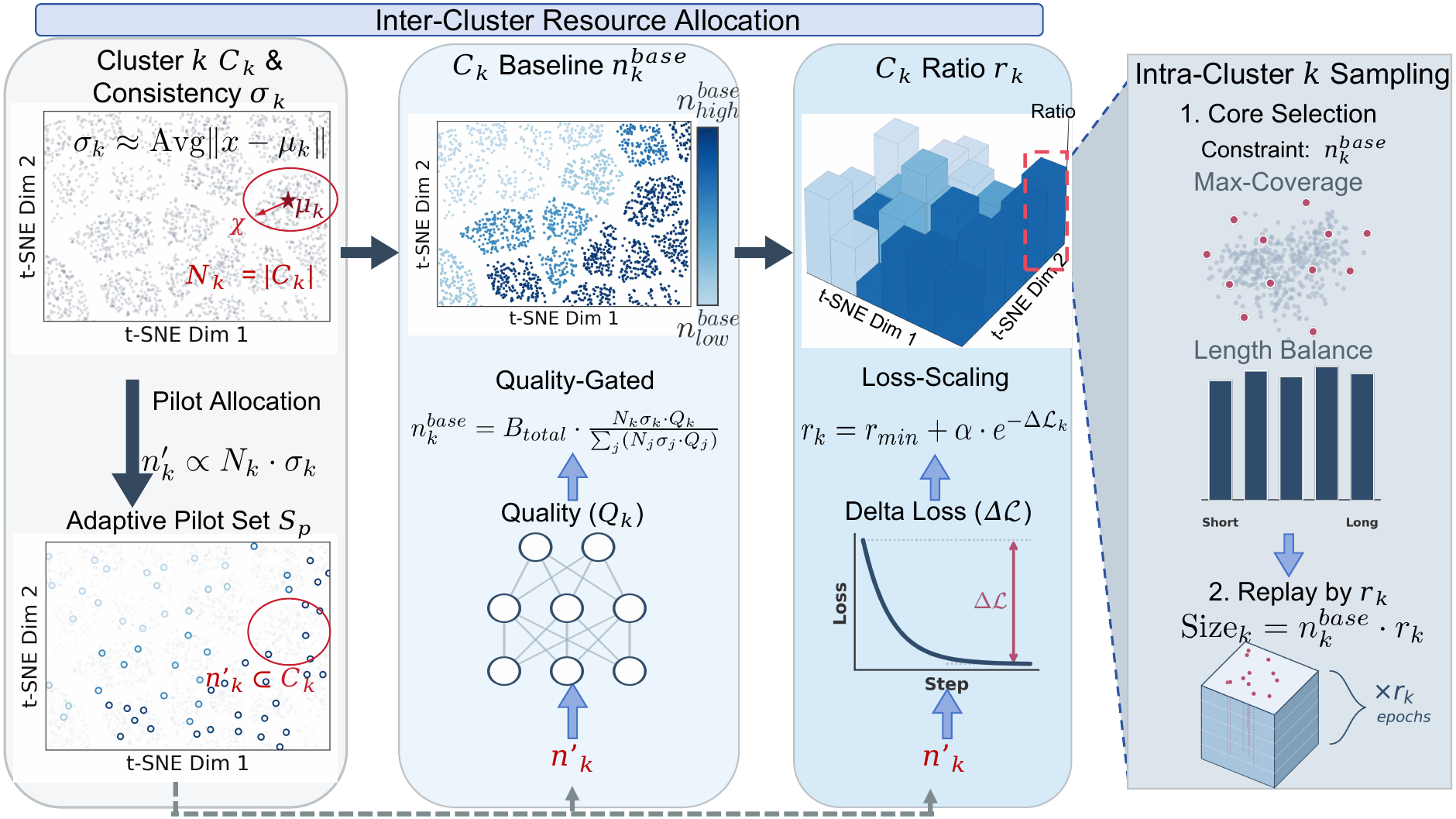}
\caption{\textbf{Overview of the GRIP Framework.} GRIP unifies \textbf{Inter-Cluster Budgeting} and \textbf{Intra-Cluster Selection} through a hierarchical geometric optimization:
\textbf{(1) Geometric Probing:} We partition the corpus into semantic clusters and construct a Neyman-optimal probe set $\mathcal{P}$ based on \textbf{Geometric Consistency} ($\sigma_k$) to estimate the baseline budget $n_k^{base}$.
\textbf{(2) Dynamic Allocation:} By monitoring the \textbf{Adaptation Delta} ($\Delta \mathcal{L}_k$), GRIP identifies representation deficits and dynamically re-allocates resources from saturated regions to high-potential clusters via a replay multiplier $r_k$.
 \textbf{(3) Rectified Sampling:} Within clusters, we employ density-based selection with a \textbf{Length-Rectification Term} to counteract embedding collapse, preserving both global structural variance and long-tail logical sequences \cite{sorscher2022beyond, ethayarajh2019contextual}.}
\label{fig:framework}
\end{figure*}

We validate GRIP by training 8B and 16B Mixture-of-Experts (MoE) models from scratch, spanning a \textbf{training scale from 100B to 300B tokens} on a hybrid corpus. Our results demonstrate that GRIP consistently achieves superior data efficiency and robustness across code generation and reasoning benchmarks compared to state-of-the-art baselines. Our main contributions are:

\begin{itemize}[leftmargin=*]

    \item \textbf{Unified Selection Framework}: We introduce \textbf{GRIP}, a hierarchical framework that unifies macro-budgeting with micro-instance selection. Evaluations on MoE architectures up to 300B tokens show that GRIP delivers a significant \textbf{+4.6\% average improvement} across benchmarks, surpassing the performance of models trained on $3\times$ larger uncurated corpora.

    \item \textbf{Adaptive Information Potential}: We propose the \textbf{Rapid Adaptation Probe (RAP)}, a mechanism grounded in $\mathcal{V}$-usable information theory. RAP identifies "representation deficits" within the geometric space, enabling \textbf{dynamic re-allocation} of the sampling budget based on the model's evolving epistemic state.

    \item \textbf{Length-Rectified Geometric Selection}: We characterize the \textbf{length-induced geometric collapse} in Transformer embeddings and introduce a rectified sampling strategy. This counteracts the suppression of long-context sequences, isolating structurally critical patterns from high-density noise.

    \item \textbf{Loss-Driven Quality Dynamics}: We establish a theoretical link between \textbf{instantaneous loss reduction} and \textbf{data learnability}. By utilizing training dynamics to reflect intrinsic data quality, our framework prioritizes samples that offer maximum incremental gain throughout the pre-training trajectory.
\end{itemize}

\section{Methodology}
\label{sec:method}

GRIP reformulates data selection as a hierarchical optimization problem over a \textbf{structured semantic landscape}, aiming to maximize cumulative information gain within a fixed computational budget. As illustrated in Figure \ref{fig:framework}, our framework operates across two coupled scales: 
First, \textbf{Inter-Cluster Budgeting} (Sec. \ref{sec:inter_cluster_budgeting}) dynamically calibrates resource allocation by synthesizing intrinsic data quality with the model's instantaneous learnability. 
Second, \textbf{Intra-Cluster Selection} (Sec. \ref{sec:intra_cluster}) employs a length-rectified geometric prior to extract high-value samples, ensuring diverse representation while explicitly counteracting embedding collapse.

\subsection{Problem Formulation}
\label{sec:problem_formulation}

Let $\mathcal{D}=\{x_i\}_{i=1}^N$ be a large corpus for language-model pretraining. Given a global sampling budget $B_{total} \ll N$, our goal is to extract a support set $S\subset\mathcal{D}$ with $|S|=B_{total}$, such that the information carried by $S$ approximates the full corpus.
We measure information through geometry in a representation space: an encoder $f_\theta$ maps each sample to a normalized embedding $\mathbf{e}(x)=f_\theta(x)\in\mathbb{S}^{d-1}$. We employ Rao's quadratic entropy (RQE) \citep{botta2005rao} as the diversity functional, a standard metric in active learning and core-set selection \citep{sener2017active}:
\begin{equation}
\mathcal{I}(\pi) \triangleq \sum_{x, y \in \mathrm{supp}(\pi)} \pi(x)\,\pi(y)\,d\big(\mathbf{e}(x), \mathbf{e}(y)\big),
\end{equation}
where $\pi$ is the sampling distribution. Our optimization objective is to find a sparse distribution $\pi^\star$ that minimizes the divergence from the corpus distribution $\pi_{\mathcal{D}}$:
\begin{equation}
\pi^\star \;=\;\arg\min_{\pi:\,|\mathrm{supp}(\pi)|=B_{total}}\; \big|\mathcal{I}(\pi)-\mathcal{I}(\pi_{\mathcal{D}})\big|.
\label{eq:diversity_match}
\end{equation}
We implement this optimization strictly within a hierarchical framework. While matching scalar entropy globally is insufficient to guarantee distributional alignment, applying this criterion locally within the semantic clusters (defined in Sec.~\ref{sec:probe_repr_cluster}) ensures that the selected subset preserves the local structural variance of the domain, thereby enforcing representational completeness across the entire geometric landscape \citep{xie2023data}.

\subsection{Probe-Centric Representation}
\label{sec:probe_repr_cluster}

\begin{figure}
    \centering
    \includegraphics[width=1\linewidth]{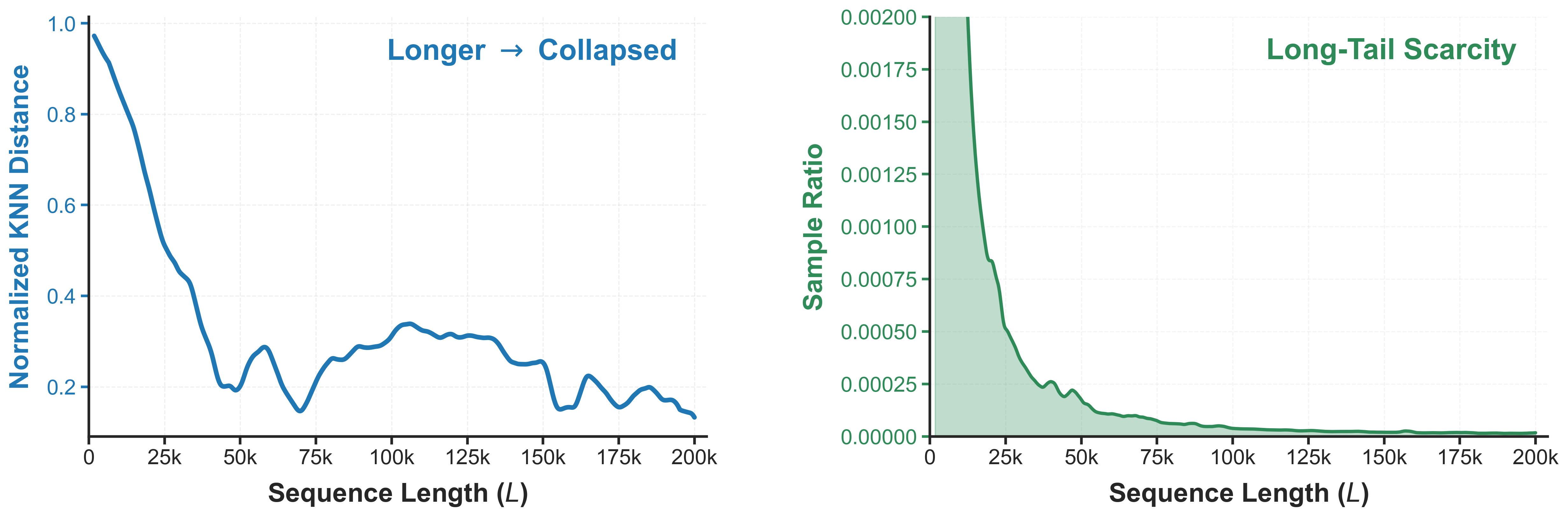}
    \caption{\textbf{Analysis of Sequence Length Pathology.} \textbf{(Left)} The normalized distance to the $k$-nearest neighbors ($k=10$) decreases rapidly as sequence length grows, indicating that embeddings of long sequences collapse into a narrow, dense region (anisotropic cone). \textbf{(Right)} The sample ratio reveals a severe heavy-tailed imbalance: data density vanishes for long sequences (power-law tail), leading to insufficient supervision for extended contexts.}
    \label{fig:length}
\end{figure}

\paragraph{Representation Space and Geometric Metrics.}
To structure the continuous semantic space of $\mathcal{D}$, we first map each sequence $x$ to a normalized embedding $\mathbf{e}(x) \in \mathbb{S}^{d-1}$ via mean-pooling over the final hidden states of the encoder $f_\theta$. Building upon this representation, we partition the space into $K$ disjoint semantic clusters $\mathcal{C} = \{C_1, \dots, C_K\}$ utilizing spherical $k$-means \citep{johnson2019billion}. 
Besides, we characterize the geometric property of each cluster $C_k$ by its \textbf{Geometric Consistency} $\sigma_k$, defined as the root mean square deviation from the centroid $\mu_k$:
\begin{equation}
\sigma_k = \sqrt{|C_k|^{-1} \sum_{x \in C_k} \|\mathbf{e}(x) - \mu_k\|^2}.
\label{eq:consistency}
\end{equation}
$\sigma_k$ quantifies Intra-Cluster Coherence: low values indicate dense semantic alignment \citep{gao2021simcse}, while high values signal dispersed distributions requiring larger sampling budgets to capture information.

\paragraph{Neyman-Optimal Probe Construction.}
To efficiently estimate cluster-level properties (e.g., quality scores $Q_k$ and training dynamics), we construct a lightweight \textbf{Probe Set} $\mathcal{P}$.
Instead of uniform sampling, we employ \textbf{Neyman’s Optimal Allocation} to minimize the variance of global estimators. We assign the probe budget $B_{probe}$ proportional to the cluster's size $N_k$ and its geometric dispersion $\sigma_k$:
\begin{equation}
n_k^{probe} \propto N_k \cdot \sigma_k, \quad \text{s.t.} \sum n_k^{probe} = B_{probe}.
\label{eq:neyman_probe}
\end{equation}
This ensures the probe prioritizes regions with high uncertainty, guaranteeing robust estimation for the subsequent budgeting stage.

\paragraph{The Length-Collapse Pathology.}
We observe a critical \textbf{Length-Induced Embedding Collapse} (Figure \ref{fig:length}). As sequences grow, embeddings collapse into a narrow cone, resulting in misleadingly high cosine similarities (pseudo-density) ~\cite{ethayarajh2019contextual,zhou2025length}. This structural failure, compounded by the scarcity of long sequences in web corpora \citep{lozhkov2024stackv2}, necessitates the length-aware rectification strategies proposed in Section \ref{sec:intra_cluster}.

\subsection{Inter-Cluster Budgeting}
\label{sec:inter_cluster_budgeting}

We formulate data selection as a two-stage allocation problem, distinguishing between \textbf{Intrinsic Quality} (static signal-to-noise ratio) and \textbf{Instantaneous Learnability} (dynamic information gain). We decouple the allocation into a static \textit{Baseline Budget} and a dynamic \textit{Replay Multiplier}.

\paragraph{Baseline Budget: Static Information Potential.}
We first allocate the global budget $B_{total}$ to maximize static information coverage. Drawing on \textbf{Neural Scaling Laws} \citep{kaplan2020scaling, sorscher2022beyond}, which suggest marginal information gain scales as a power law of data mass, we propose a \textbf{Non-Linear Capacity Allocation} rule:

\begin{equation}
n_k^{base} = B_{total} \cdot \frac{(N_k \cdot \sigma_k)^\tau \cdot \exp(Q_k / T)}{\sum_{j=1}^K (N_j \cdot \sigma_j)^\tau \cdot \exp(Q_j / T)}.
\label{eq:baseline_budget}
\end{equation}

Here, the sub-linear exponent $\tau < 1$ prevents resource monopolization by massive clusters \citep{hoffmann2022training}, while the Boltzmann term $\exp(Q_k / T)$ acts as a soft spectral filter, prioritizing high-signal regions ($Q_k$) estimated by the probe.

\begin{figure}
    \centering
    \includegraphics[width=0.8\linewidth]{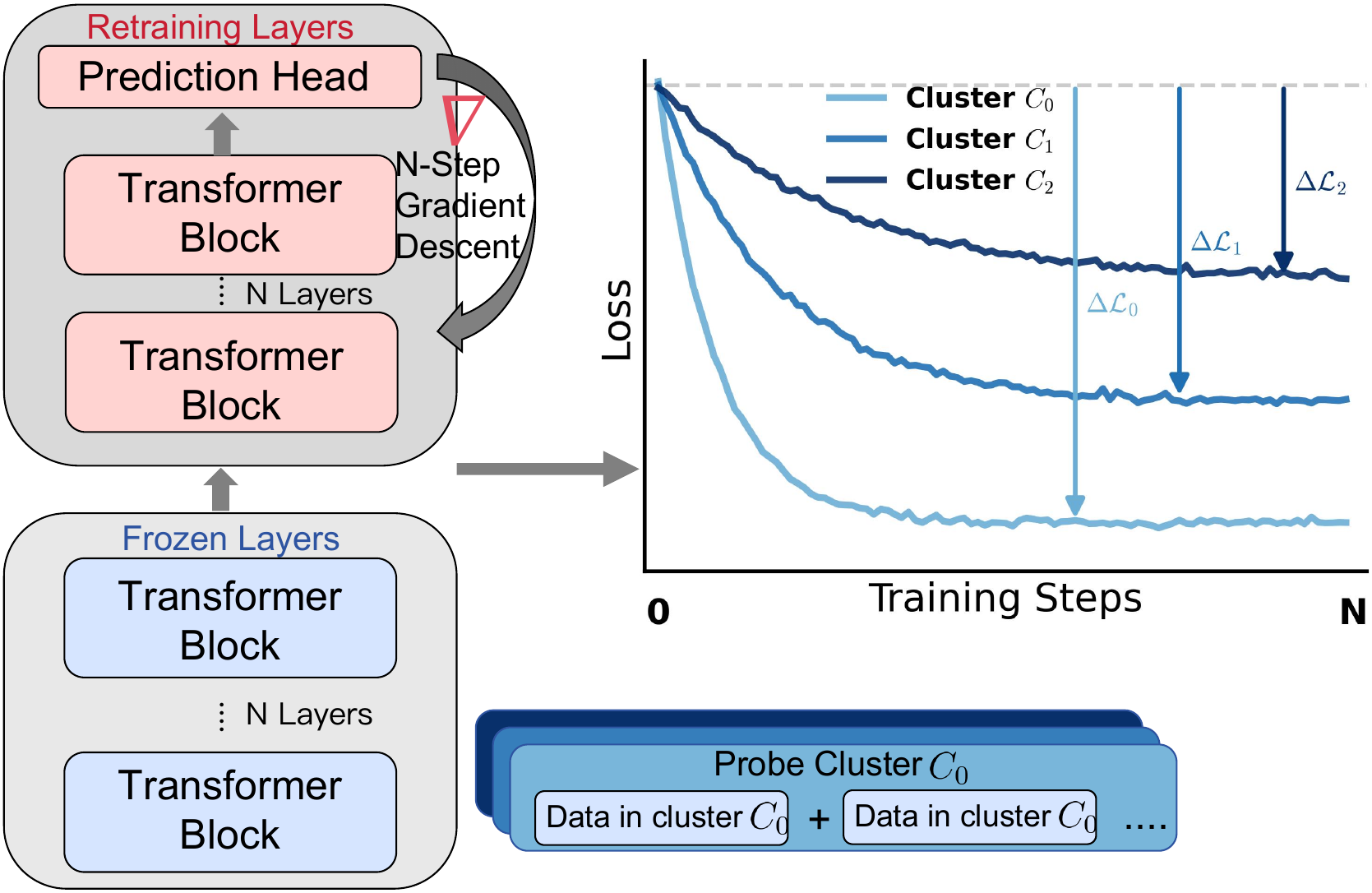}
    \caption{\textbf{Mechanism of the Rapid Adaptation Probe.} To evaluate the learnability of different clusters (e.g., $C_0, C_1, C_2$), we freeze the lower layers and \textbf{reset the Retraining Layers to a consistent initialization}. We then perform $N$-step gradient descent independently for each cluster. The resulting \textbf{Adaptation Delta} ($\Delta \mathcal{L}_k$) measures how quickly the loss drops from this common starting point. A rapid loss reduction (e.g., $C_0$) indicates that the data is easily predictable given the current features, implying low incremental information gain, while a small drop (e.g., $C_2$) indicates a learning bottleneck requiring increased replay budget.}
    \label{fig:loss_dynamics}
\end{figure}

\paragraph{Closed-Loop Replay via Loss Dynamics.}
To align allocation with evolving model capabilities, we leverage the framework of \textbf{$\mathcal{V}$-Usable Information} \citep{xu2020theory, ethayarajh2022understanding}. We define the utility of cluster $C_k$ as its \textbf{Instantaneous Information Potential}—the reduction in entropy achievable given the current feature representation.
As illustrated in Figure~\ref{fig:loss_dynamics}, we estimate this via a Rapid Adaptation Probe. We partition the model into \textbf{Frozen Layers} (lower transformer blocks) and Retraining Layers (upper transformer blocks and prediction head). 
To ensure a standardized comparison, for each cluster $C_k$, we reset the Retraining Layers to a shared initialization $\theta_{init}$ before performing $N$-step gradient descent. We then measure the Adaptation Delta:
\begin{equation}
\Delta \mathcal{L}_k = \mathcal{L}(\theta_{init}; C_k) - \mathcal{L}(\theta_{N}; C_k).
\label{eq:deltaloss}
\end{equation}
Here, $\mathcal{L}(\theta; C_k)$ denotes the loss on cluster $C_k$ given parameters $\theta$.
Starting from the same initialization, a rapid loss reduction (large $\Delta \mathcal{L}_k$, e.g., $C_0$) indicates that the data is easily predictable with current features, implying low incremental information gain. Conversely, a small drop ($\Delta \mathcal{L}_k \to 0$) signals a \textbf{Representation Deficit}, where the model struggles to acquire features. We target these bottlenecks by setting the replay multiplier $r_k$ inversely proportional to the drop:
\begin{equation}
r_k = 1 + \alpha \cdot \exp\left(-\frac{\Delta \mathcal{L}_k}{\tau_{norm}}\right) \cdot \mathbb{I}(Q_k > \tau_{th}).
\label{eq:replay_dynamics}
\end{equation}
To standardize the learnability signal across varying sequence lengths, we normalize the adaptation delta by its population expectation $\tau_{norm} = \mathbb{E}_{k}[\Delta \mathcal{L}_k]$. This normalization decouples the metric from scale discrepancies, ensuring it reflects relative information gain. 
Based on this, the replay intensity is derived inversely to the relative drop, directing the budget toward clusters where the model's current representation is insufficient. 
To further refine this objective, we distinguish between valid learnable deficits (epistemic uncertainty) and irreducible noise (aleatoric uncertainty) by strictly coupling the replay intensity with a quality gate $\mathbb{I}(Q_k > \tau_{th})$. 
This integration ensures that the sampling budget is concentrated on data that is both information-rich and currently under-fitted \citep{qin2023infobatch}.

\paragraph{Global Budget Normalization.}
We synthesize the static capacity and dynamic learnability into a unified \textbf{Effective Sampling Utility}, defined as $\mathcal{U}_k^{(t)} = n_k^{base} \cdot r_k^{(t)}$.
To adhere to the strict computational constraint $B_{total}$ defined in Eq. \ref{eq:diversity_match}, we formulate the final allocation using a global partition function $Z^{(t)}$:
\begin{equation}
n_k^{(t)} = \frac{B_{total}}{Z^{(t)}} \cdot \mathcal{U}_k^{(t)}, \quad \text{where } Z^{(t)} = \sum_{j=1}^K \mathcal{U}_j^{(t)}.
\label{eq:global_norm}
\end{equation}
This renormalization implements a \textbf{Zero-Sum Resource Redistribution} \citep{xie2023doremi}, automatically siphoning budget from saturated clusters to fund high-deficit regions.


\subsection{Intra-Cluster Selection}
\label{sec:intra_cluster}

Having established the macro-level budget $n_k^{(t)}$ for each cluster, our final objective is to select specific instances that maximize local geometric coverage.
We approach this as a dual-objective sampling problem designed to mitigate semantic redundancy while explicitly counteracting the length-bias pathology inherent in embedding spaces.

\paragraph{Kernel-Based Diversity Sampling.}
Within a semantic cluster $C_k$, data points are typically non-uniformly distributed. To enforce \textbf{Local Geometric Diversity}, we employ an \textbf{Inverse Propensity Sampling} strategy~\cite{zhang2025harnessing}. We model the local density $\rho(x)$ of a sample $x$ using a Gaussian Kernel over its nearest neighbors:
\begin{equation}
\rho(x) = \sum_{z \in \mathcal{N}_k(x)} \exp\left(-\frac{\|\mathbf{e}(x) - \mathbf{e}(z)\|^2}{2h^2}\right).
\end{equation}
We define the base sampling probability as $P_{base}(x) \propto \rho(x)^{-1}$.
This mechanism actively penalizes samples located in the dense centroids of the cluster—which typically correspond to common, low-surprisal patterns \citep{abbas2023semdedup}—and promotes the selection of distinct examples that define the convex hull of the cluster's semantic span \citep{tirumala2023d4}.

\paragraph{Length-Rectified Importance Weighting.}
Standard density-based sampling fails for long-context data due to the \textbf{Length-Induced Collapse} (Sec. \ref{sec:probe_repr_cluster}).
Since long sequences collapse into a narrow cone, they exhibit artificially high cosine similarities (pseudo-density), causing naive samplers to discard them.
To correct this geometric distortion, we introduce a \textbf{Length-Rectification Term} $\beta$:
\begin{equation}
P_{select}(x) \propto \underbrace{\frac{1}{\rho(x)}}_{\text{Diversity}} \cdot \underbrace{\left(\frac{\ell(x)}{\ell_{avg}}\right)^\beta}_{\text{Length Compensation}}.
\label{eq:length_rectified}
\end{equation}

Here, $\beta \ge 1$ acts as a restoration coefficient. By up-weighting long sequences, we effectively "re-expand" the collapsed embedding cone \citep{li2020sentence}.
The final subset $S_k$ is drawn from $C_k$ using $P_{select}$ without replacement until the target size $n_k^{(t)}$ is met \citep{bai2024longalign}.

\section{Experimental Setup}
\label{sec:exp_setup}

\begin{table*}[t]
\centering
\setlength{\tabcolsep}{4pt}
\renewcommand{\arraystretch}{1.25}
\caption{\textbf{Main Results and Ablation Study.} Comparisons on code generation, reasoning, and multilingual benchmarks. \textbf{Panel A} shows scaling efficiency at 300B tokens; \textbf{Panel B} details the progressive ablation at 100B tokens. We report \textbf{Pass@1} for generation tasks.}
\label{tab:main_results}
\resizebox{\linewidth}{!}{%
    \begin{tabular}{l c | c c c c c c c c}
    \toprule
    \rowcolor{headergray}
     & \textbf{Avg.} & \multicolumn{2}{c}{\textbf{HumanEval}} & \multicolumn{2}{c}{\textbf{MBPP}} & \textbf{LiveCode} & \multicolumn{2}{c}{\textbf{CruxEval}} & \textbf{MultiPL-E} \\
    \rowcolor{headergray}
    \textbf{Model \& Method} & \textbf{Score} & \textbf{Pass@1} & \textbf{HE$^+$} & \textbf{Pass@1} & \textbf{MBPP$^+$} & \textbf{Pass@1} & \textbf{Input} & \textbf{Output} & \textbf{Avg.} \\
    \midrule
    \multicolumn{10}{l}{\textit{\textbf{Panel A: Scaling Performance (Total Budget: 300B tokens = 100B tokens $\times$ 3 Epochs)}}} \\
    \midrule
    \noalign{\gdef\BaseEight{20.2}} 
    Random 8B & \num{20.2} & 38.5 & 36.2 & 27.4 & 30.1 & 1.2 & 16.5 & 18.8 & 13.5 \\
    \rowcolor{highlightyellow}
    \textbf{GRIP 8B} & \Score{\BaseEight}{24.80} & \Score{38.5}{52.36} & \Score{36.2}{46.02} & \Score{27.4}{36.02} & \Score{30.1}{42.59} & \Score{1.2}{5.27} & \Score{16.5}{24.12} & \Score{18.8}{25.75} & \Score{13.5}{23.67} \\
    \midrule[0.2pt]
    \noalign{\gdef\BaseSixteen{20.8}}
    Random 16B & \num{20.8} & 40.8 & 38.5 & 29.1 & 32.5 & 1.5 & 18.2 & 20.4 & 14.8 \\
    \rowcolor{highlightyellow}
    \textbf{GRIP 16B} & \Score{\BaseSixteen}{25.62} & \Score{40.8}{55.36} & \Score{38.5}{48.22} & \Score{29.1}{38.54} & \Score{32.5}{44.81} & \Score{1.5}{6.43} & \Score{18.2}{26.25} & \Score{20.4}{26.88} & \Score{14.8}{25.47} \\
    \midrule
    \multicolumn{10}{l}{\textit{\textbf{Panel B: Ablation (GRIP-8B, Total Budget: 100B tokens)}}} \\
    \midrule
    1. Random & \num{18.7} & 35.2 & 32.1 & 24.5 & 28.1 & 0.8 & 12.5 & 15.2 & 11.3 \\
    2. + Static Budget & \num{19.4} & 38.6 & 35.4 & 26.2 & 30.4 & 1.3 & 14.4 & 17.6 & 13.1 \\
    3. + Static Replay & \num{20.2} & 41.2 & 38.5 & 28.1 & 32.2 & 1.8 & 16.3 & 18.5 & 14.2 \\
    4. + Loss Replay & \num{21.2} & 44.8 & 41.5 & 30.8 & 34.9 & 2.4 & 19.1 & 20.8 & 16.5 \\
    5. + Diversity & \num{21.3} & 45.4 & 42.0 & 30.5 & 34.5 & 2.5 & 19.3 & 20.9 & 16.0 \\
    \rowcolor{highlightyellow}
    \textbf{6. GRIP (Full)} & \Score{18.7}{22.0} & \Score{35.2}{47.2} & \Score{32.1}{43.8} & \Score{24.5}{32.5} & \Score{28.1}{38.2} & \Score{0.8}{2.9} & \Score{12.5}{20.5} & \Score{15.2}{21.8} & \Score{11.3}{19.2} \\
    \bottomrule
    \end{tabular}%
}
\end{table*}


We evaluate \textbf{GRIP} on \textbf{large-scale code pre-training}. Code is a demanding setting for data curation: small syntax issues can break execution \citep{li2023starcoder}, the corpus spans many programming languages \citep{feng2020codebert}, and programs follow hierarchical dependencies \citep{guo2020graphcodebert}. These properties make naive filtering brittle, allowing us to strictly test whether GRIP preserves both geometric coverage and local structural integrity.

\subsection{Corpus and Model Configuration}
\label{sec:corpus_model}

\paragraph{Hybrid Corpus Composition.}
To simulate realistic domain augmentation, we construct a \textbf{100B token} hybrid candidate pool $\mathcal{D}_{pool}$ comprising a \textbf{Fixed Background} (CommonCrawl) and a \textbf{Selectable Foreground} (The Stack v2 \citep{lozhkov2024stackv2}). 
We employ the \textit{Qwen3 embedding model}~\cite{zhang2025qwen3} to map all candidate documents into a unified dense vector space. Our selection framework operates exclusively on The Stack; the curated high-value subset is then combined with the constant CommonCrawl data to train models from scratch.

\paragraph{Model Architecture.}
We adopt a fine-grained sparse Mixture-of-Experts (MoE) Transformer, increasing the total expert capacity while keeping the number of activated parameters per token fixed to preserve inference efficiency~\citep{deepseek_v2}. We instantiate \textbf{8B} (32 experts) and \textbf{16B} (64 experts), both with \textbf{1.4B} active parameters. All models are trained \textbf{from scratch} with the same training recipe; only the data curation differs

\begin{figure}[t]
    \centering
    \includegraphics[width=1\linewidth]{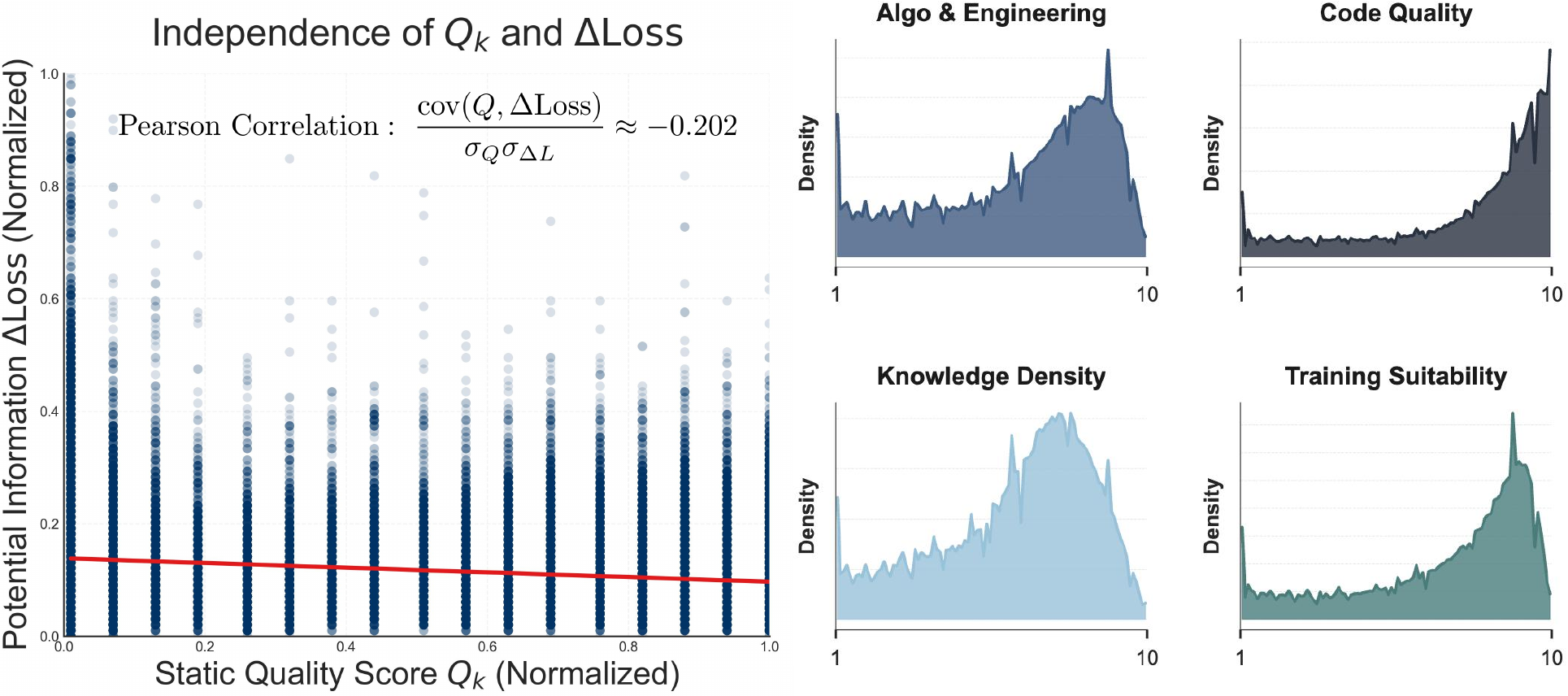}
    \caption{\textbf{Independence of Static Quality and Training Dynamics.} \textbf{(Left)} Scatter plot where each data point represents a cluster, showing a weak correlation (Pearson $\approx -0.202$) between LLM-based quality scores $Q_k$ and the adaptation delta $\Delta \mathcal{L}_k$. \textbf{(Right)} Density distributions across four dimensions illustrate that high-value semantic features exhibit distinct spectral signatures.}
    \label{fig:quality_independence}
\end{figure}

\subsection{GRIP Implementation Details}
\label{sec:implementation}

\paragraph{Probe Construction and Static Quality.}
We construct a Neyman-optimal probe $\mathcal{P}$ ($|\mathcal{P}| \approx 0.5\%$) to serve as a statistical proxy for the corpus. 
We derive cluster quality $Q_k$ by applying the \textbf{LLM-as-a-Judge} paradigm \citep{zheng2024judging} directly to the probe samples using \texttt{Qwen3-235B} \citep{qwen3report}. 
Documents are scored on educational value and correctness to formulate the aggregate metric $Q_k$. Exact prompts are detailed in Appendix \ref{app:quality_prompt}.

\paragraph{Look-Ahead Dynamics and Loss Calculation.}
To strictly measure the \textit{learnability} of data clusters rather than mere memorization, we implement a \textbf{Parameter Reset Protocol}.
For a given probe batch from cluster $C_k$:

\begin{itemize}[leftmargin=*, noitemsep, topsep=2pt]
    \item \textbf{Proxy Model Selection:} We utilize \textbf{SmolLM-135M} and \textbf{360M} as efficient proxies for look-ahead dynamics, while \textbf{Qwen-2.5} serves as a larger reference to validate the consistency of learnability rankings.
    \item \textbf{Partial Reset:} We freeze the backbone and re-initialize the weights of the LM head and the last $N$ transformer layers to a random state. We configure the reset depth $N \in \{0, 1, 2, 4\}$.
    \item \textbf{Adaptation Step:} We optimize the reset parameters using AdamW (lr=$3\times 10^{-4}$) and perform $K=10$ training epochs over the cluster-specific batch to ensure sufficient feature adaptation.
    \item \textbf{Metric Computation:} To decouple the metric from sequence length variations, we normalize the loss by the token count $L$ (length-normalization). We then compute the learnability score as the relative reduction: $\delta \mathcal{L} = (\mathcal{L}_{init} - \mathcal{L}_{final}) / \mathcal{L}_{init}$.
\end{itemize}
This metric isolates the gradient information gain available to the model given its current representation.

\begin{figure}[t]
    \centering
    \includegraphics[width=1\linewidth]{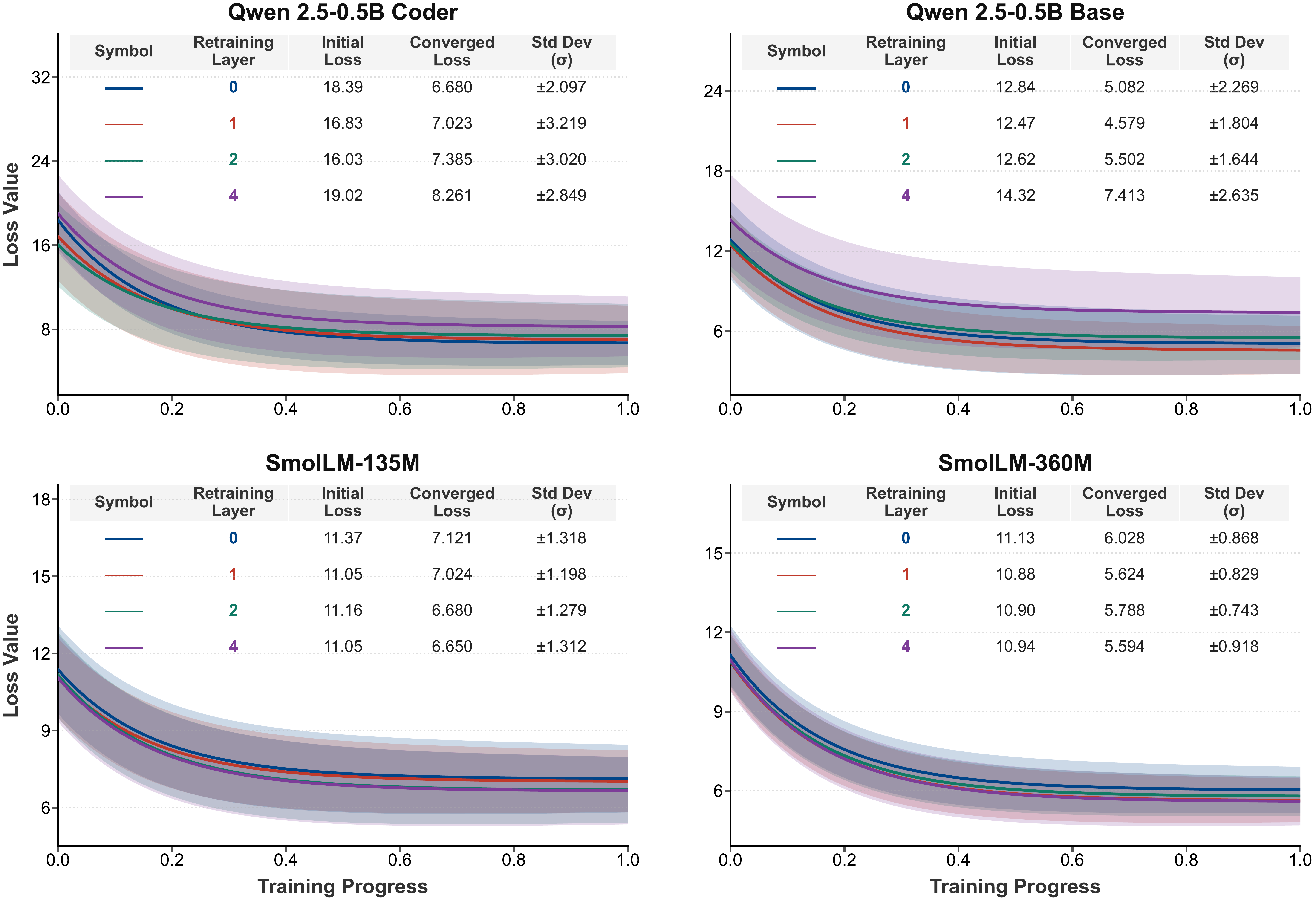}
    \caption{\textbf{Cross-Model Transferability of Loss Dynamics.} Loss trajectories across different model families (Qwen-2.5 vs. SmolLM) and reset depths ($N \in \{0, 1, 2, 4\}$). The high consistency in ranking suggests that lightweight proxy models can effectively guide data selection for larger target architectures.}
    \label{fig:proxy_consistency}
\end{figure}

\paragraph{Hyperparameters.}
We set the probe gradient steps $N=10$ to efficiently capture the loss convergence \textit{elbow}. In \textbf{Dynamic Replay} (Eq.~\ref{eq:replay_dynamics}), we set the intensity $\alpha=2.0$ (allowing a max $3\times$ budget). Crucially, to address varying loss magnitudes caused by factors like sequence length, we set the normalization factor $\tau_{norm} \approx \mathbb{E}[\Delta \mathcal{L}]$. This standardizes the adaptation delta, ensuring the replay mechanism responds to relative learnability rather than absolute scale. For \textbf{Inter-Cluster Budgeting} (Eq.~\ref{eq:baseline_budget}), we adopt square-root sampling ($\tau=0.5$) to ensure sub-linear scaling, with temperature $T=1.0$ to preserve the natural quality distribution. Finally, we set the length-penalty $\beta=0.3$ in \textbf{Intra-Cluster Selection} to mitigate embedding collapse. Experiments are conducted on 8$\times$ NVIDIA H800 GPUs using AdamW.

\subsection{Evaluation Protocol}

\paragraph{Baselines and Ablations.}
We design a progressive ablation path to verify each component:
\begin{itemize}
    \item \textbf{Random Sampling (Lower Bound):} Uniformly samples tokens from The Stack without any filtering, serving as the baseline for data distribution.
    \item \textbf{Static Quality Budgeting:} Applies only the \textit{Baseline Budget} ($n_k^{base}$, Eq. \ref{eq:baseline_budget}) based on probe-estimated quality $Q_k$. This variant ignores training dynamics (Replay) and intra-cluster geometry.
    \item \textbf{Static + Quality-Based Replay:} Introduces a replay mechanism, but calculates the multiplier $r_k$ using static quality scores ($Q_k$) rather than loss dynamics. This baseline tests whether the \textit{dynamic} nature of our feedback loop is necessary, or if simply up-weighting high-quality static data suffices.
    \item \textbf{Static + Loss-Based Replay (No Diversity):} Incorporates our proposed \textit{Closed-Loop Replay} (Eq. \ref{eq:replay_dynamics}) based on $\delta \mathcal{L}$, but employs random sampling within clusters. This isolates the gain from the macro-level budgeting strategy.
    \item \textbf{Static + Loss-Based Replay + Diversity (No Length Fix):} Adds the \textit{Kernel-Based Diversity Sampling} (Eq. \ref{eq:length_rectified}, first term) but omits the length-rectification term ($\beta=0$). This validates the specific contribution of our length-aware correction.
    \item \textbf{GRIP (Ours):} The full framework integrating Inter-Cluster Budgeting, Loss-Based Replay, and Length-Rectified Intra-Cluster Selection.
\end{itemize}

\paragraph{Benchmarks.}
We evaluate performance across three dimensions:
\begin{itemize}[leftmargin=*, noitemsep, topsep=2pt]
    \item \textbf{Code Generation:} We utilize \textit{HumanEval(+)} (\textbf{HE}) \citep{chen2021evaluating} and \textit{MBPP(+)} \citep{austin2021program} to evaluate generation performance.
    \item \textbf{Robustness \& Reasoning:} We assess temporal generalization using \textit{LiveCodeBench} (\textbf{LCB}) \citep{jain2024livecodebench} and execution prediction capabilities using \textit{CruxEval} \citep{gu2024cruxeval}.
    \item \textbf{Multilingual Proficiency:} We conduct a fine-grained analysis across multiple programming languages using \textit{MultiPL-E} \citep{cassano2023multiple}.
    
\end{itemize}


\section{Result Analysis}
\label{sec:results}

\subsection{Scaling Efficiency and Generalization}
\label{sec:result_scaling}

We evaluate the performance of GRIP across two model scales (8B and 16B) to assess its scaling efficiency. As shown in \textbf{Table \ref{tab:main_results} (Panel A)}, GRIP consistently outperforms the random sampling baseline, with the performance gap widening as model capacity increases.

\textbf{Widening Gains at Scale.} GRIP achieves a substantial average score improvement of \textbf{+4.6\%} on the 8B model, which further expands to \textbf{+4.8\%} on the 16B architecture. This trend indicates that while random selection suffers from diminishing returns due to noise accumulation, our geometric and learnability-driven strategy effectively saturates the larger model's capacity with high-information density data, preventing the "data exhaustion" typically seen in scaling.

\textbf{Superior Reasoning and Robustness.} The advantages of GRIP are most pronounced in reasoning-intensive benchmarks, specifically \textbf{LiveCodeBench} (+4.1\% on 8B) and \textbf{MultiPL-E} (+10.2\% on 8B). Unlike standard benchmarks that may rely on rote memorization of common patterns, these results demonstrate that GRIP prioritizes data containing complex logical structures and diverse syntax, thereby significantly enhancing the model's ability to generalize across languages and solve novel algorithmic challenges.

\subsection{Progressive Ablation Study}
\label{sec:result_ablation}

To verify the contribution of each component, we analyze the ablation path in \textbf{Table \ref{tab:main_results} (Panel B)}.

\begin{itemize}[leftmargin=*]

    \item \textbf{Efficacy of Static Optimization (Row 1--3):} The transition from Random to \textbf{Static Budgeting} (Row 2) delivers a solid initial boost (+0.7\% Avg), validating our probe-based strategy for optimizing macro-distributions. Subsequently, adding \textbf{Static Replay} (Row 3) yields a further +0.8\% gain. These results establish that while prioritizing high-quality ($Q_k$) data provides a strong foundation, purely static filtering lacks the adaptability to target the model's evolving information needs.

    \item \textbf{Static vs. Dynamic Replay (Row 3 vs. 4):} Transitioning from static quality budgeting to our \textbf{Loss-Based Replay} yields a consistent improvement (\textbf{+1.0\%} Avg). This confirms that static heuristics are insufficient for capturing the evolving needs of the model; instead, the instantaneous \textit{learnability} signal ($\Delta \mathcal{L}$) effectively identifies and up-weights "hard-to-learn" clusters that harbor the highest marginal information gain.

    \item \textbf{The "Diversity Trap" (Row 4 vs. 5):} Crucially, applying \textbf{Kernel-Based Diversity Sampling} alone results in stagnation (\textbf{+0.1\%} Avg) and a notable performance drop in structural tasks like \textbf{MultiPL-E} ($16.5 \to 16.0$). This reveals a "diversity trap": naive density sampling falls victim to the embedding collapse pathology, erroneously discarding high-value long-context code as redundant, which harms the model's ability to maintain long-range dependencies.

    \item \textbf{Length Rectification (Row 5 vs. 6):} The full GRIP framework, incorporating \textbf{Length Rectification}, resolves this pathology and delivers a decisive final boost (\textbf{+0.7\%} Avg). By explicitly "re-expanding" the probability mass for collapsed long sequences, we observe a robust recovery in multilingual and reasoning capabilities (e.g., \textbf{MultiPL-E} rebounds to 19.2). This validates that geometric diversity is only effective when rectified for the structural biases of the Transformer's representation space.
\end{itemize}

\subsection{Validation of Selection Dynamics and Geometric Analysis}
\label{sec:result_proxy}

\paragraph{Independence of Static Quality and Loss Dynamics.}
As illustrated in Figure~\ref{fig:quality_independence}, we observe a weak correlation (Pearson $\approx -0.202$) between static quality $Q_k$ and dynamic learnability $\Delta \mathcal{L}_k$ across semantic clusters. Notably, clusters with near-zero static quality exhibit a wide range of potential information gain, suggesting that even ostensibly "low-quality" data can harbor significant learning utility depending on the model's current state. Furthermore, the density profiles across dimensions—\textit{Algo \& Engineering}, \textit{Code Quality}, \textit{Knowledge Density}, and \textit{Training Suitability}—highlight the multi-faceted distribution of high-value data. This independence underscores that static heuristics alone cannot fully capture data importance; integrating dynamic loss feedback is essential to identify incremental information gain relative to the model's evolving representation.

\paragraph{Robustness and Transferability of Proxy-Guided Selection.}
To strictly decouple intrinsic data learnability from model-specific memorization capacity, we conduct a comprehensive sensitivity analysis using four proxy architectures (SmolLM-135M/360M and Qwen-2.5-0.5B Base/Coder) across varying parameter reset depths $N \in \{0, 1, 2, 4\}$, where $N=0$ denotes retraining only the prediction head.
As illustrated in Figure~\ref{fig:proxy_consistency}, while stronger architectures like Qwen-Coder exhibit heightened sensitivity to data quality—manifesting in larger adaptation deltas ($\Delta \mathcal{L} \approx 11$) and wider discriminative variance ($\sigma \approx 2.8$) compared to smaller models (e.g., SmolLM-360M: $\Delta \mathcal{L} \approx 5, \sigma \approx 0.8$)—the structural behavior remains remarkably consistent.
Crucially, the relative ranking of cluster utility proves robust across both reset depth and model scale: the convergence topologies of ultra-lightweight proxies mirror those of larger reference models, confirming that the learnability signal is an intrinsic geometric property of the data rather than an artifact of training capacity. 
This dual consistency validates our deployment of the computationally efficient $N=0$ protocol, which faithfully guides the data budget for the 16B target model with a negligible overhead of less than 1\% of total pre-training FLOPs.

\paragraph{Geometric Collapse and Length Rectification.}
Figure~\ref{fig:length} provides empirical confirmation of the embedding pathology targeted by our sampling strategy. The left panel reveals a precipitous decay in normalized $k$-NN distance—dropping below $0.2$ for sequences exceeding $50\text{k}$ tokens—while the right panel highlights the extreme scarcity of such samples in the natural distribution. This intersection creates a "double jeopardy": rare long sequences exhibit artificially high density, causing standard diversity samplers to erroneously discard them as redundant. By implementing length-rectified weighting to counteract this suppression, GRIP ensures adequate supervision for extended contexts, directly contributing to the significant gains observed in complex reasoning benchmarks (e.g., +5.7\% on MultiPL-E) where maintaining long-range structural dependencies is critical.

\section{Conclusion}
\label{sec:conclusion}
In this work, we presented \textbf{GRIP}, a framework that maximizes data efficiency by treating pre-training as a dynamic semantic space approximation problem. By integrating a rapid adaptation probe to resolve macro-level redundancy and length-rectified sampling to counteract micro-level geometric collapse, GRIP aligns data allocation with the model's evolving learning state. Our results on 8B and 16B MoE models demonstrate that prioritizing informative geometry over raw volume enables state-of-the-art performance under fixed computational constraints, providing a scalable path for efficient large-scale curation.


\bibliography{iclr2026_conference}
\bibliographystyle{iclr2026_conference}

\appendix
\section{Appendix}

\subsection{Prompts for Static Quality Assessment}
\label{app:quality_prompt}

To ensure the selected code data meets rigorous pre-training standards, we employ an \textbf{LLM-as-a-Judge} paradigm. Unlike generic scoring, we utilize a specialized "Senior Technical Auditor" persona designed to penalize non-functional code while rewarding pedagogical value and engineering robustness. The full system instruction is provided below.

\begin{AuditProtocol}
\textbf{\large System Instruction: Comprehensive Code Audit Protocol}

You are acting as a \textbf{Senior Technical Auditor}. Your objective is to rigorously evaluate the provided code snippet against four core performance indicators. 

Unlike simple correctness checks, you must assess the code's readiness for high-quality pre-training datasets. Use a granular \textbf{0 to 10 scale} for each dimension.

\vspace{0.3cm}
\hrule
\vspace{0.3cm}

\textbf{1. Code Quality \& Compliance (0-10)} \\
\textit{Focus: Syntax, naming, and readability.}
\begin{itemize}[leftmargin=15pt, nosep, labelsep=5pt]
    \item \textbf{0-3 (Low):} Severe syntax errors; code cannot run; messy formatting or naming makes it unreadable.
    \item \textbf{4-7 (Mid):} Executable and generally correct; minor formatting issues or inconsistent naming; readable but not polished.
    \item \textbf{8-10 (High):} Error-free and standard compliant (e.g., PEP8); precise naming; clean and professional structure.
\end{itemize}

\vspace{0.3cm}

\textbf{2. Algorithmic \& Engineering Design (0-10)} \\
\textit{Focus: Modularity, robustness, and structure.}
\begin{itemize}[leftmargin=15pt, nosep, labelsep=5pt]
    \item \textbf{0-3 (Low):} All code in one block (global scope); no functions/classes; lacks input validation or error handling.
    \item \textbf{4-7 (Mid):} Split into functions; basic error checking; logic is structured but lacks extensibility.
    \item \textbf{8-10 (High):} High modularity (classes/functions); robust error handling; follows best engineering practices.
\end{itemize}

\vspace{0.3cm}

\textbf{3. Training Suitability (0-10)} \\
\textit{Focus: Educational value and clarity for learning.}
\begin{itemize}[leftmargin=15pt, nosep, labelsep=5pt]
    \item \textbf{0-3 (Low):} No comments; confusing logic; uses "magic numbers"; hard for a model or human to learn from.
    \item \textbf{4-7 (Mid):} Clear logic; includes basic comments for key steps; suitable for general training.
    \item \textbf{8-10 (High):} Excellent comments explaining the "why"; demonstrates best practices; highly suitable for fine-tuning/teaching.
\end{itemize}

\vspace{0.3cm}

\textbf{4. Knowledge Density (0-10)} \\
\textit{Focus: Technical insight and complexity.}
\begin{itemize}[leftmargin=15pt, nosep, labelsep=5pt]
    \item \textbf{0-3 (Low):} Trivial code (e.g., simple print statements, basic loops); contains little domain knowledge.
    \item \textbf{4-7 (Mid):} Standard algorithms or typical business logic; useful but not unique.
    \item \textbf{8-10 (High):} complex algorithms, deep optimizations, or cross-domain knowledge; high technical value.
\end{itemize}

\vspace{0.3cm}
\hrule
\vspace{0.3cm}

\textbf{Response Requirement:}
Analyze the input code and output a JSON object strictly adhering to this schema:

\begin{verbatim}
{
  "syntax_integrity_score": <int 0-10>,
  "architecture_score": <int 0-10>,
  "educational_value_score": <int 0-10>,
  "technical_depth_score": <int 0-10>
}
\end{verbatim}

\textbf{Target Code Snippet:}
\$content

\end{AuditProtocol}

\end{document}

%% file: math_commands.tex
\usepackage{amsmath,amsfonts,bm}

\def\eqref#1{equation~\ref{#1}}

\def\1{\bm{1}}

\DeclareMathAlphabet{\mathsfit}{\encodingdefault}{\sfdefault}{m}{sl}
\SetMathAlphabet{\mathsfit}{bold}{\encodingdefault}{\sfdefault}{bx}{n}

